\documentclass[11pt]{article}

\usepackage[preprint]{acl}

\usepackage{times}
\usepackage{latexsym}

\usepackage[T1]{fontenc}

\usepackage[utf8]{inputenc}

\usepackage{microtype}

\usepackage{inconsolata}

\usepackage{graphicx}

\usepackage{multirow}
\usepackage{booktabs}
\usepackage{makecell}
\usepackage{xcolor}
\usepackage{vcell}
\usepackage{colortbl}
\usepackage{amssymb}
\usepackage{pifont}
\usepackage{adjustbox}
\usepackage{amsmath}
\usepackage{tabularx}
\usepackage[toc]{multitoc}
\usepackage[skins]{tcolorbox}
\usepackage{soul}
\usepackage{subcaption}
\usepackage{fvextra}
\usepackage{enumitem}
\usepackage{xspace}
\usepackage{url}

\definecolor{wkred}{RGB}{240, 190, 190}
\definecolor{wkblue}{RGB}{190, 210, 235}
\definecolor{wkgreen}{RGB}{190, 225, 200}
\definecolor{wkpurple}{RGB}{210,210,253}
\definecolor{wkyellow}{RGB}{255,241,177}
\definecolor{wkgold}{RGB}{255, 223, 129}
\definecolor{wksilver}{RGB}{192, 192, 192}

\definecolor{codegreen}{rgb}{0,0.6,0}
\definecolor{codegray}{rgb}{0.5,0.5,0.5}
\definecolor{codepurple}{rgb}{0.58,0,0.82}
\definecolor{backcolour}{rgb}{0.95,0.95,0.92}

\definecolor{upred}{HTML}{DC143C}
\definecolor{downgreen}{HTML}{32CD32}

\newcommand{\code}[1]{\nolinkurl{#1}}

\newcommand{\up}[1]{\textcolor{upred}{{+#1}}}

\newcommand{\best}{\cellcolor{wkred}}
\newcommand{\second}{\cellcolor{wkblue}}

\newcommand{\frameworkname}{MathCanvas\xspace}
\newcommand{\editname}{MathCanvas-Edit\xspace}
\newcommand{\imagenname}{MathCanvas-Imagen\xspace}
\newcommand{\datasetname}{MathCanvas-Instruct\xspace}
\newcommand{\benchmarkname}{MathCanvas-Bench\xspace}
\newcommand{\modelname}{BAGEL-Canvas\xspace}

\title{MathCanvas: Intrinsic Visual Chain-of-Thought for Multimodal Mathematical Reasoning}

\author{
 \textbf{Weikang Shi\textsuperscript{1}\thanks{Equal Contribution}}
 \textbf{Aldrich Yu\textsuperscript{1}\footnotemark[1]}
 \textbf{Rongyao Fang\textsuperscript{1}\footnotemark[1]\thanks{Project lead}}
 \textbf{Houxing Ren\textsuperscript{1}}
 \textbf{Ke Wang\textsuperscript{1}}
 \textbf{Aojun Zhou\textsuperscript{1}}
 \textbf{Changyao Tian\textsuperscript{1}}
\\
 \textbf{Xinyu Fu\textsuperscript{2}}
 \textbf{Yuxuan Hu\textsuperscript{1}}
 \textbf{Zimu Lu\textsuperscript{1}}
 \textbf{Linjiang Huang\textsuperscript{3}}
 \textbf{Si Liu\textsuperscript{3}}
 \textbf{Rui Liu\textsuperscript{2}\thanks{Corresponding author}}
 \textbf{Hongsheng Li\textsuperscript{1}\footnotemark[3]}
\\
\\
 \textsuperscript{1}Multimedia Laboratory (MMLab), The Chinese University of Hong Kong,
\\
 \textsuperscript{2}Huawei Research,
 \textsuperscript{3}BUAA
\\
 \small{
   \href{mailto:wkshi@link.cuhk.edu.hk}{wkshi@link.cuhk.edu.hk} $\quad$
   \href{mailto:hsli@ee.cuhk.edu.hk}{hsli@ee.cuhk.edu.hk}
 }
}

\makeatletter
\AtBeginDocument{
\let\@oldmaketitle\@maketitle
\renewcommand{\@maketitle}{\@oldmaketitle
  \vspace{-5pt}
  \includegraphics[width=\linewidth]{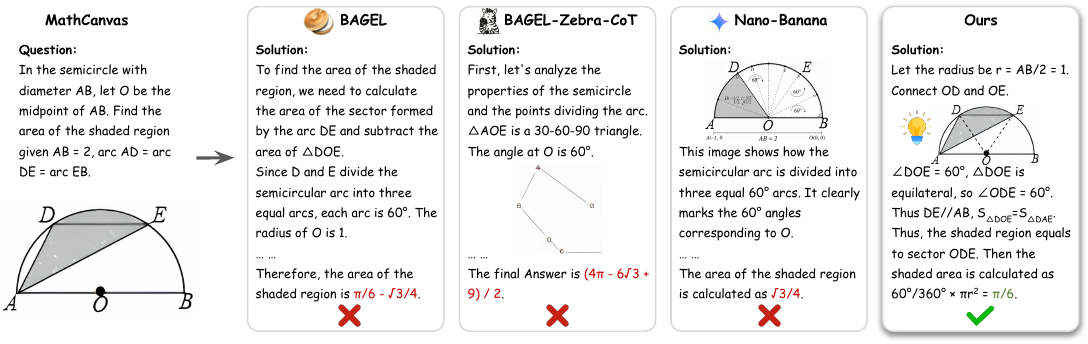}
  \vspace{-18pt}
  \captionof{figure}{
    MathCanvas demonstrates the first successful application of intrinsic Visual Chain-of-Thought (VCoT) for complex mathematical reasoning. Prior attempts fail by generating incorrect (BAGEL-Zebra-CoT) or strategically poor (Nano-Banana) visuals, leading to wrong solutions. In contrast, MathCanvas correctly generates an intermediate visual step that unlocks a simpler, elegant solution path.
  }
  \label{fig:teaser}
  \vspace{17pt}
 }
}
\makeatother

\begin{document}
\maketitle

\begin{abstract}
While Large Language Models (LLMs) have excelled in textual reasoning, they struggle with mathematical domains like geometry that intrinsically rely on visual aids. Existing approaches to Visual Chain-of-Thought (VCoT) are often limited by rigid external tools or fail to generate the high-fidelity, strategically-timed diagrams necessary for complex problem-solving. 
To bridge this gap, we introduce \frameworkname, a comprehensive framework designed to endow unified Large Multimodal Models (LMMs) with intrinsic VCoT capabilities for mathematics. Our approach consists of two phases. First, a \textit{Visual Manipulation} stage pre-trains the model on a novel 15.2M-pair corpus, comprising 10M caption-to-diagram pairs (\imagenname) and 5.2M step-by-step editing trajectories (\editname), to master diagram generation and editing. Second, a \textit{Strategic Visual-Aided Reasoning} stage fine-tunes the model on \datasetname, a new 219K-example dataset of interleaved visual-textual reasoning paths, teaching it \textit{when} and \textit{how} to leverage visual aids. To facilitate rigorous evaluation, we introduce \benchmarkname, a challenging benchmark with 3K problems that require models to produce interleaved visual-textual solutions. Our model, \modelname, trained under this framework, achieves an 86\% relative improvement over strong LMM baselines on \benchmarkname, demonstrating excellent generalization to other public math benchmarks. Our work provides a complete toolkit—framework, datasets, and benchmark—to unlock complex, human-like visual-aided reasoning in LMMs. Project Page: \url{https://mathcanvas.github.io/}

\end{abstract}

\section{Introduction}

Mathematical reasoning represents a pinnacle of human intelligence, demanding a sophisticated interplay of logical deduction, symbolic manipulation, and abstract thinking. The advent of Large Language Models (LLMs)~\citep{ deepseekai2025deepseekr1incentivizingreasoningcapability, yang2024qwen25mathtechnicalreportmathematical, openai2024openaio1card} has marked a significant milestone in artificial intelligence, demonstrating remarkable capabilities in tackling complex mathematical reasoning tasks. 
A key driver of recent progress in LLM-based reasoning has been the Chain-of-Thought (CoT)~\citep{wei2023chainofthoughtpromptingelicitsreasoning} technique, which enables models to externalize intermediate steps and significantly improves performance on mathematical tasks.

However, the purely textual nature of CoT presents a fundamental limitation in domains like geometry and function analysis, where human problem-solving intrinsically involves constructing and manipulating visual aids, and even state-of-the-art models struggle in its absence (see Figure~\ref{fig:teaser_LMM} in Appendix~\ref{app:reasoning_results}).
This gap has motivated the development of Visual Chain-of-Thought (VCoT), which aims to integrate visual information into the reasoning process. Early approaches to VCoT have predominantly relied on external specialized tools, such as dedicated vision models~\citep{shao2024visual, hu2024visualsketchpadsketchingvisual, gao2025interleavedmodalchainofthought} or code interpreters~\citep{hu2024visualsketchpadsketchingvisual, wang2025visuothinkempoweringlvlmreasoning, wang2025geometryzeroimprovinggeometrysolving}. While effective in specific contexts, these tool-based methods are often rigid, constrained to a predefined set of operations, and dependent on specific input formats (e.g., source code), which hinders their flexibility and broader applicability. 
Recent work has explored intrinsic VCoT, where unified large multimodal models (LMMs) natively generate visual thoughts as an integral part of their reasoning process~\citep{cheng2025visualthoughtsunifiedperspective, li2025imagine, li2025zebra, chern2025thinking}.

Though promising, these previous attempts have been confined to simple domains and have yet to succeed in mathematics due to two key challenges. First, current unified LMMs lack the capability to generate and iteratively edit the high-fidelity mathematical diagrams required for precise reasoning. The generated visuals are often geometrically incorrect, rendering them useless for logical deduction, as shown with BAGEL-Zebra-CoT~\citep{li2025zebra} in Figure~\ref{fig:teaser}. Second, and more fundamentally, models lack the procedural knowledge to employ visual aids as a strategic component of their reasoning process—the complex decision of determining \emph{when} to draw, \emph{what} to draw, and \emph{how} to leverage the visualization for subsequent logical deduction. This strategic failure is evident even in advanced models like Nano-Banana~\citep{comanici2025gemini25pushingfrontier}, shown in Figure~\ref{fig:teaser}, whose generated visual acts more as a flawed decoration than an integral reasoning step, ultimately failing to uncover the key insight needed for the solution.

To this end, we argue that addressing these challenges requires models capable of interleaving textual deduction with the creation and modification of visual aids. Accordingly, we introduce \textbf{\frameworkname}, a comprehensive framework designed to endow unified LMMs with intrinsic VCoT capabilities for complex mathematical problem-solving. 
Our approach is structured around two complementary phases: \emph{Visual Manipulation} and \emph{Strategic Visual-Aided Reasoning}.

The first phase, \emph{Visual Manipulation}, focuses on equipping the model with foundational visual synthesis and editing skills. To achieve this, we construct a new million-scale pretraining corpus specifically for mathematical diagrams. This resource comprises two parts: \editname, containing 5.2M step-by-step diagram editing instruction pairs generated via a hybrid pipeline that combines LLM-driven mining with programmatic synthesis, and \imagenname, with 10M caption-to-diagram pairs. Pretraining on them imparts the robust diagram generation and manipulation abilities that form the bedrock of our approach.

The second phase, \emph{Strategic Visual-Aided Reasoning}, aims to teach the model how to interleave diagrammatic actions with its textual reasoning steps. For this purpose, we curate \textbf{\datasetname}, the first large-scale dataset for interleaved visual–textual mathematical reasoning. It contains 219K training examples, where each solution is represented as an interleaved sequence of textual reasoning and corresponding visual steps. As demonstrated in Figure~\ref{fig:teaser}, training on \datasetname enables the model to learn how to coordinate diagrammatic actions with reasoning trajectories to successfully solve complex problems.

Furthermore, to rigorously evaluate models’ capabilities in visual–textual mathematical reasoning, we introduce a dedicated benchmark test set \textbf{\benchmarkname} comprising 3K carefully curated problems. Each test instance requires the solver to produce coherent interleaved reasoning and visual outputs. We benchmarked 20 leading LMMs on this dataset, revealing substantial performance gaps and establishing it as a challenging and comprehensive testbed for future research on Visual Chain-of-Thought reasoning.

\begin{figure*}[t]
 \centering
 \includegraphics[width=1.0\linewidth]{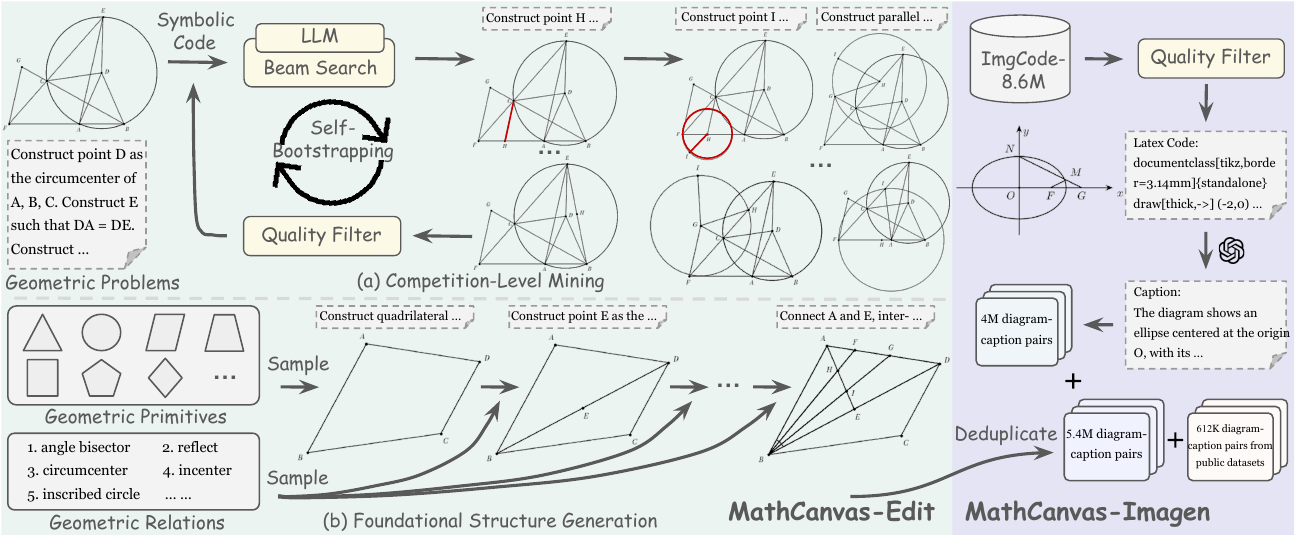}
 \caption{The curation pipeline for the \editname and \imagenname dataset.}
 \label{fig:data_pipeline}
 \vspace{-1mm}
\end{figure*}

In summary, our contributions are as follows:
\begin{itemize}[leftmargin=5mm, itemsep=0mm]
    \item We propose \frameworkname, a comprehensive framework that enables LMMs to perform \textbf{intrinsic VCoT} reasoning for complex mathematical problem solving.
    \item We construct two large-scale corpora tailored for our two-phase approach: a 15.2M-pair pretraining dataset for \emph{Visual Manipulation}, and a 219K-example fine-tuning dataset for \emph{Strategic Visual-Aided Reasoning}.
    \item We further introduce a dedicated \benchmarkname test set with 3K problems and benchmark 20 leading LMMs on it, revealing substantial deficiencies and establishing a challenging evaluation bed for future research.
    \item Experiments show that our model trained under the \frameworkname framework achieves a \textbf{86\% relative improvement} over strong LMM baselines on \benchmarkname, demonstrating the effectiveness of our approach in unlocking intrinsic VCoT capabilities.
\end{itemize}

\section{Related Work}

\paragraph{Mathematical Reasoning with Large Multimodal Models.}
The remarkable success of text-only LLMs in mathematical reasoning, often driven by sophisticated chain-of-thought prompting~\citep{wei2023chainofthoughtpromptingelicitsreasoning,yang2024qwen25mathtechnicalreportmathematical,yue2023mammothbuildingmathgeneralist,wang2023mathcoderseamlesscodeintegration,shao2024deepseekmathpushinglimitsmathematical}, has naturally spurred interest in extending these capabilities to the multimodal domain. Initial efforts in this area have largely involved adapting LMMs by enhancing vision-text alignment on domain-specific data and then fine-tuning on mathematical question-answer pairs~\citep{gao2025gllavasolvinggeometricproblem, wang2025mathcodervlbridgingvisioncode,zhuang2024mathpumaprogressiveupwardmultimodal,zhang2024mavismathematicalvisualinstruction,guo2025mammothvlelicitingmultimodalreasoning}. While subsequent work has advanced the state of the art with techniques like reinforcement learning~\citep{yang2025kwaikeyevl15technical,wang2025internvl35advancingopensourcemultimodal,duan2025got,wei2025advancingmultimodalreasoningreinforcement}, these models remain fundamentally text-centric.  While they effectively interpret visual information in the input, they largely neglect vision as an active, generative component of the reasoning process itself.

\paragraph{Visual Chain-of-Thought.}
Unlike various textual chain-of-thought~\citep{wei2023chainofthoughtpromptingelicitsreasoning,fang2025got,fang2025flux}, visual chain-of-thought aims to bridge this gap by integrating the generation of visual aids directly into the reasoning process. Existing approaches follow two main lines. The first leverages external tools, such as vision models to extract image details~\citep{shao2024visual, chen2025mint,hu2024visualsketchpadsketchingvisual,openai2025o3systemcard,gao2025interleavedmodalchainofthought} or code interpreters to add auxiliary structures~\citep{hu2024visualsketchpadsketchingvisual, wang2025visuothinkempoweringlvlmreasoning, wang2025geometryzeroimprovinggeometrysolving}. This approach, however, is constrained, as these tools are either non-generative or lack general applicability due to rigidity. The second line explores intrinsic VCoT, where models natively generate visual thoughts as an integral part of their reasoning~\citep{cheng2025visualthoughtsunifiedperspective, li2025imagine, li2025imaginereasoningspacemultimodal, chern2025thinking, li2025zebra}. Despite its promise, this approach has so far been demonstrated primarily in simpler domains like spatial games and struggles to produce the precise, logically consistent diagrams required for complex mathematical reasoning.

\paragraph{Datasets and Benchmarks for Multimodal Mathematical Reasoning.}
The progress in visual-mathematical reasoning is largely driven by the evolution of its benchmarks. While foundational datasets like Geometry3K~\citep{lu2021intergpsinterpretablegeometryproblem} and ScienceQA~\citep{lu2022learn} established the task, recent challenging benchmarks such as MMMU~\citep{yue2023mmmu}, MathVista~\citep{lu2024mathvista}, Mathvision~\citep{wang2024measuring}, and MathVerse~\citep{zhang2024mathverse}, among others~\citep{qiao2024wemathdoeslargemultimodal,wang2025benchmarkingmultimodalmathematicalreasoning,sun2024mmmathadvancingmultimodalmath}, have pushed the limits of LMMs' visual reasoning. However, a fundamental limitation persists: these benchmarks consist of static problem-solution pairs and lack the step-by-step \textit{visual} demonstrations required to train models for dynamic, process-oriented reasoning. This is precisely the gap our work addresses with the introduction of \datasetname and the \benchmarkname benchmark.

\section{Method}

In this section, we detail the methodology behind \frameworkname. We first describe the construction of our large-scale training corpora for visual manipulation and strategic reasoning (\ref{sec:trainingdataset}). We then introduce \benchmarkname, a dedicated benchmark for rigorous evaluation (\ref{sec:benchmark}). Finally, we present our two-stage training recipe that leverages these resources to instill intrinsic VCoT capabilities in a unified LMM (\ref{sec:recipe}).

\subsection{Training Corpora Construction}
\label{sec:trainingdataset}

\subsubsection{Million-scale Pretraining Corpus}
To endow unified LMMs with the foundational visual synthesis and editing capabilities required for mathematical reasoning, we construct a comprehensive million-scale pretraining corpus comprising two complementary components: \editname for diagram editing and \imagenname for diagram generation. The overall construction pipeline is shown in Figure~\ref{fig:data_pipeline}.

\paragraph{\editname} is designed to teach models how to iteratively modify mathematical diagrams through step-by-step transformations. We construct this dataset through a hybrid pipeline that combines complex competition-level geometry problems with systematically generated simple geometric figures, yielding a total of 5.2M edit trajectories.

\noindent\textbf{Competition-Level Mining.} We start with 128 geometry problems from mathematical competitions to serve as realistic seed configurations. Using these seeds, we employ the AlphaGeometry LLM~\citep{AlphaGeometryTrinh2024} with beam search to generate numerous auxiliary line drawing methods for each problem. We then filter for geometrically invalid constructions and render the corresponding diagram sequences, where each step is an edit operation (e.g., adding an auxiliary line, marking an angle). This iterative process yields 4.2M edit trajectories capturing the complexity of competition-level reasoning. To ensure visual diversity from this limited set of seeds, the rendering of each trajectory is controlled by a unique random seed, varying visual attributes like orientation and line styles.

\noindent\textbf{Foundational Structure Generation.} While competition problems provide realism, they tend toward complexity that may not adequately cover fundamental editing operations. To address this, we construct a complementary set of simple geometric figures using AlphaGeometry's formal language. We first define a basic geometric primitive set (e.g., points, lines, circles) and a geometric relation set (e.g., circumcenter, incenter, parallel), the full details of which are provided in Appendix~\ref{app:pretrain_details}. Then we develop an automated algorithm that randomly and incrementally adds geometric primitives and relations to these basic structures, creating progressively more complex diagrams. Invalid or degenerate configurations are filtered out through geometric constraint checking. By leveraging different random seeds during rendering, we obtain 1M additional edit trajectories that provide systematic coverage of fundamental geometric operations after three iterations of this synthetic generation process.

\begin{figure*}[t]
 \centering
 \includegraphics[width=1.0\linewidth]{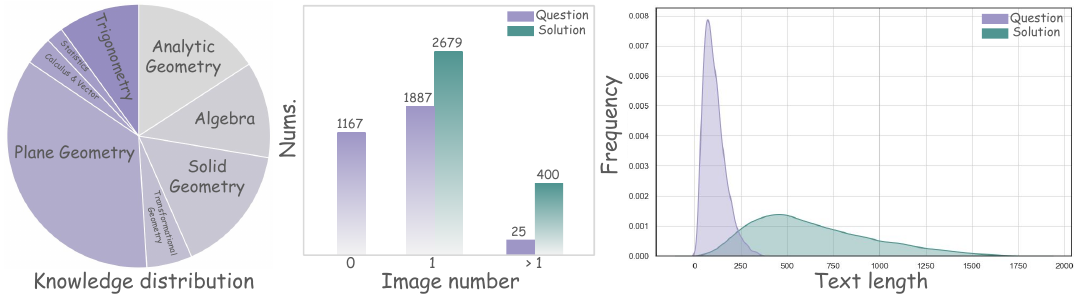}
 \caption{Statistical analysis of the \benchmarkname test set. \textbf{Left:} Knowledge types distribution. \textbf{Middle:} Distribution of questions and solutions containing varying numbers of images. \textbf{Right:} Text length distribution of questions and solutions (measured in text tokens).}
 \label{fig:bench}
 \vspace{-1mm}
\end{figure*}

\paragraph{\imagenname} focuses on teaching models to generate mathematical diagrams from textual descriptions. We construct it by aggregating and processing data from three complementary sources, resulting in 10M caption-to-diagram pairs.

\noindent\textbf{Re-purposing from \editname.} We first leverage the edit trajectories in \editname, extracting caption-to-diagram pairs from each editing step. After deduplication based on visual and textual similarity, we obtain 5.4M diverse caption-to-diagram pairs that inherently align with the types of diagrams needed for mathematical reasoning.

\noindent\textbf{Augmenting with Code-derived Captions.} To further scale our dataset, we utilize the ImgCode-8.6M~\citep{wang2025mathcodervlbridgingvisioncode} dataset, which contains programmatically generated mathematical diagrams paired with source code. We first apply quality filtering to remove corrupted or low-quality images. We then employ GPT-4.1-mini to generate natural language captions by taking image-code pairs as input, producing descriptions that capture both the visual content and mathematical semantics of each diagram. This process yields 4M high-quality caption-to-diagram pairs with rich, descriptive captions with diverse mathematical diagrams.

\noindent\textbf{Incorporating Public Datasets.} Finally, we incorporate 612K caption-to-diagram pairs from existing public resources, including MAVIS~\citep{zhang2024mavismathematicalvisualinstruction} and TR-CoT~\citep{deng2025theoremvalidatedreversechainofthoughtproblem}, which provide additional diversity in caption styles and diagram types, complementing our dataset.

Through this comprehensive construction process, the pretraining corpus provides a robust foundation for pretraining models on both diagram generation and editing, establishing the essential visual capabilities needed for intrinsic VCoT in mathematical reasoning.

\subsubsection{\datasetname}
\label{sec:instruct_dataset}

To equip models with the ability to strategically interleave visual synthesis and editing actions with their textual reasoning process, we introduce \datasetname, the first large-scale dataset specifically designed for interleaved visual-textual mathematical reasoning.

\paragraph{Dataset Construction} We begin by gathering 632K multimodal mathematics problems and solutions from a wide array of middle school and high school textbooks, exams, and websites. From this initial pool, we implement a rigorous multi-stage filtering pipeline to ensure data quality and relevance. First, we employ GPT-5 to analyze the problems, filtering out examples where the provided images served no role in the reasoning process. This step also standardized all mathematical formulas into LaTeX format, resulting in a refined set of 367K problems. A second round of filtering, also powered by GPT-5, removes problems that contained errors, lacked explicit answers, featured low-quality or unclear images, or consisted solely of drawing tasks. This left us with 303K high-quality problems. 

To ensure the novelty and diversity of the dataset, we then perform both text and image deduplication, which yielded 222K unique problem-solution pairs. The images in the remaining dataset underwent a quality enhancement step using a super-resolution model, SwinIR~\citep{liang2021swinirimagerestorationusing}, to improve clarity and detail before being resized to a uniform 512x512 resolution. Finally, GPT-4.1 is used to classify all problems into a hierarchical taxonomy of 8 major categories and fine-grained subcategories.
This collection is then partitioned to form our evaluation benchmark, \benchmarkname{}, with the remaining 219K examples constituting the \datasetname{} training set. Further statistics and examples for \datasetname{} are presented in Appendix~\ref{app:sft_details}.

\begin{figure*}[t]
 \centering
 \includegraphics[width=1.0\linewidth]{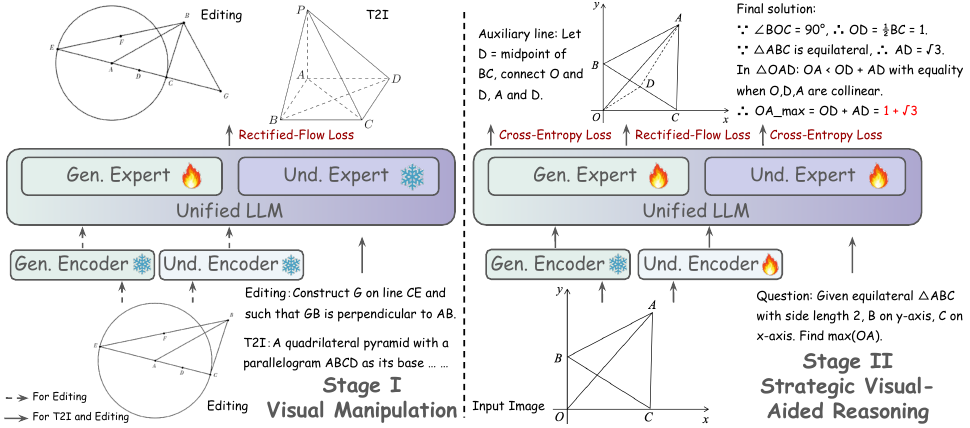}
 \caption{The two-stage training recipe for MathCanvas. \textbf{(Left) Stage I: Visual Manipulation.} The model's Generation Expert is pretrained on our \editname{} and \imagenname{} corpora to instill foundational diagram generation and editing skills. 
\textbf{(Right) Stage II: Strategic Visual-Aided Reasoning.} The entire model is then fine-tuned on \datasetname{} to learn the strategic interleaving of visual actions with textual reasoning.}
 \label{fig:model}
 \vspace{-1mm}
\end{figure*}

\subsection{The \benchmarkname{} Evaluation Benchmark}
\label{sec:benchmark}

\paragraph{Benchmark Construction} We construct \benchmarkname{} by sampling 3K problems from the 222K-pair collection described in Section~\ref{sec:instruct_dataset}.
The construction process involves three key steps. First, we exclude all multiple-choice questions to ensure that evaluation relies on generative reasoning rather than random guessing.
Second, to create a balanced test set, we perform weighted sampling across problem categories, setting the sampling weight for each category to the 0.7 exponential power of its proportion. This strategy increases the representation of less common problem types.
Finally, to prevent data leakage, we remove any question from the remaining 219K training set that has a 5-gram Jaccard similarity score higher than 0.4 with any problem in \benchmarkname{}. This process helps ensure a fair evaluation of model generalization.
Further statistics on the final benchmark are shown in Figure~\ref{fig:bench}.

\paragraph{Evaluation Protocol} Our evaluation protocol relies on GPT-4.1 to ensure consistent and scalable assessment. For each problem, GPT-4.1 is tasked with extracting the final answers for every sub-question from the model's output and comparing them against the ground-truth answers. The specific prompt templates used for this process are detailed in Appendix~\ref{app:eval_details}. We employ two distinct metrics to score performance:

\textbf{Complete Accuracy:} A binary score is awarded. A model receives 1 point only if the answers to all sub-questions are correct, and 0 otherwise. This metric evaluates the model's ability to solve a problem completely.

\textbf{Weighted Scoring:} To provide a more granular evaluation of partial progress, this metric assigns exponentially increasing weights to each sub-question. The precise formula for this weighting scheme is detailed in Appendix~\ref{app:eval_details}. The final score is the sum of the weights of the correctly answered sub-questions, a method that allows us to assess the model's accuracy on intermediate steps within the reasoning chain.

Thus, \benchmarkname{} provides a rigorous and challenging testbed for evaluating interleaved image-textual reasoning capabilities.

\begin{table*}[t]
\centering
\resizebox{\textwidth}{!}{%
\begin{tabular}{l|c|c|cc|cccccccc}
\toprule
\multirow{2.5}{*}{\textbf{Model}} & 
\multirow{2.5}{*}{\textbf{Size}} & 
\multirow{2.5}{*}{\textbf{Think}} &
\multicolumn{2}{c|}{\textbf{Overall}} &
\multirow{2.5}{*}{\textbf{Algebra}} & 
\multirow{2.5}{*}{\textbf{\parbox{1.3cm}{\centering Analytic\\Geom.}}} & 
\multirow{2.5}{*}{\textbf{\parbox{1.3cm}{\centering Calc \&\\Vector}}} & 
\multirow{2.5}{*}{\textbf{\parbox{1.3cm}{\centering Plane\\Geom.}}} & 
\multirow{2.5}{*}{\textbf{\parbox{1.3cm}{\centering Solid\\Geom.}}} & 
\multirow{2.5}{*}{\textbf{Stats.}} & 
\multirow{2.5}{*}{\textbf{\parbox{1.3cm}{\centering Transf.\\Geom.}}} &
\multirow{2.5}{*}{\textbf{Trig.}} 
\\
\cmidrule(lr){4-5}

~ & ~ & ~ &
\textbf{Complete} & \textbf{Weighted} & ~ & ~ & ~ & ~ & ~ & ~ & ~ & \\
\toprule

\multicolumn{13}{c}{\textit{Closed-source (unified) LMMs}}\\
\midrule
Gemini-2.5-Pro       & - & \ding{51} & \best{47.9} & \best{58.2} & 68.0 & \best{59.2} & 60.2 & \best{54.8} & \best{48.7} & 64.5 & \best{58.5} & \best{69.9} \\
Gemini-2.5-Flash       & - & \ding{51} & 39.3 & 49.5 & 63.2 & 56.5 & 54.6 & 40.7 & 40.7 & 61.1 & 46.8 & 64.6 \\
Gemini-2.0-Flash       & - & \ding{55} & 21.2 & 32.6 & 39.1 & 32.6 & 38.9 & 31.1 & 25.6 & 51.4 & 28.1 & 38.0 \\
GPT-4.1       & - & \ding{55} & 19.0 & 30.0 & 40.4 & 30.7 & 37.1 & 24.1 & 25.1 & 54.0 & 21.5 & 42.5 \\
GPT-4.1-mini       & - & \ding{55} & 14.6 & 26.3 & 35.7 & 30.5 & 36.5 & 22.0 & 22.4 & 24.8 & 19.7 & 30.3 \\
GPT-4o       & - & \ding{55} & 9.9 & 19.4 & 21.6 & 17.7 & 21.8 & 19.5 & 18.6 & 17.4 & 13.2 & 23.0 \\
GPT-5       & - & \ding{51} & 43.5 & 51.4 & \best{68.7} & 55.5 & \best{64.2} & 45.6 & 36.1 & 64.5 & 42.7 & 66.5 \\
Claude-Sonnet-4       & - & \ding{51} & 25.0 & 37.8 & 44.8 & 38.9 & 49.3 & 33.8 & 33.0 & 46.9 & 30.3 & 47.6 \\
Seed-1.6-Thinking       & - & \ding{51} & 44.1 & 55.2 & 67.7 & 57.5 & 55.9 & 52.2 & 45.0 & 65.1 & 56.8 & 60.7 \\
Qwen3-VL-Plus       & - & \ding{51} & 40.9 & 51.5 & 67.0 & 54.6 & 56.9 & 45.9 & 42.0 & \best{66.7} & 49.3 & 58.9 \\
Nano-Banana       & - & \ding{55} & 33.2 & 43.7 & 55.4 & 50.2 & 51.8 & 34.5 & 36.6 & 56.7 & 39.4 & 60.4 \\

\midrule

\multicolumn{13}{c}{\textit{Open-source (unified) LMMs}}\\
\midrule
Qwen-2.5-VL-7B       & 7B & \ding{55} & 8.9 & 18.7 & 19.5 & 19.0 & 19.2 & 20.6 & 18.7 & 10.7 & 13.9 & 15.0 \\
Qwen-2.5-VL-32B       & 32B & \ding{55} & 15.4 & 27.6 & 29.8 & 27.4 & 27.8 & 27.4 & 27.2 & 27.9 & 20.1 & 30.5 \\
Qwen-2.5-VL-72B       & 72B & \ding{55} & 21.1 & 32.8 & 30.6 & 19.5 & \second{36.4} & 34.5 & 33.5 & 23.9 & \second{33.6} & \second{48.9} \\
Gemma-3-27b-it       & 27B & \ding{55} & 15.8 & 26.6 & 31.3 & 28.4 & 34.4 & 25.8 & 21.0 & 40.0 & 21.0 & 26.9 \\
InternVL3.5-8B       & 8B & \ding{55} & 16.7 & 26.4 & 32.3 & \second{33.8} & 33.8 & 24.2 & 26.9 & \second{43.7} & 16.2 & 14.9 \\
InternVL3.5-30B-A3B       & 30B & \ding{55} & 11.7 & 22.2 & 22.2 & 19.9 & 15.1 & 24.9 & 24.3 & 22.1 & 17.4 & 18.4 \\
Keye-VL-1.5-8B       & 8B & \ding{51} & 17.1 & 27.0 & \second{33.1} & 28.0 & 26.2 & 27.0 & 23.6 & 29.5 & 20.9 & 26.3 \\

BAGEL       & 7B & \ding{55} & 8.3 & 18.5 & 18.1 & 13.1 & 17.1 & 20.8 & 23.0 & 10.9 & 19.4 & 13.3 \\
BAGEL‑Zebra‑CoT       & 7B & \ding{55} & 8.0 & 16.6 & 18.0 & 15.1 & 15.6 & 18.0 & 16.8 & 20.8 & 11.1 & 14.1 \\

\midrule

\textbf{\modelname} & 7B & \ding{55} & \second{21.9} & \second{34.4} & 29.9 & 27.2 & 17.9 & \second{40.0} & \second{35.3} & 23.2 & 29.3 & 40.4 \\

$\Delta$ Over Base Model & ~ & ~ & \up{13.6} & \up{15.9} & \up{11.8} & \up{14.1} & \up{0.8} & \up{19.2} & \up{12.3} & \up{12.3} & \up{9.9} & \up{27.1} \\

\bottomrule
\end{tabular}%
}
\caption{Comparison of model performances across all mathematical subjects. The best \colorbox{wkred}{closed-source} and \colorbox{wkblue}{open-source} highest accuracy of LMMs are marked in {red} and {blue}, respectively.}
\label{tab:main_model_performance}

\end{table*}

\subsection{Two-Stage Training Recipe}
\label{sec:recipe}

We implement our framework on BAGEL~\citep{deng2025emergingpropertiesunifiedmultimodal}, a state-of-the-art unified LMM. Its architecture features two distinct transformer experts—one for understanding and one for generation—integrated within a single, unified model structure. This design provides a strong foundation for our approach. Our \frameworkname~recipe enhances this architecture through a two-stage process, as illustrated in Figure~\ref{fig:model}: a foundational Stage I: Visual Manipulation, followed by Stage II: Strategic Visual-Aided Reasoning.

\paragraph{Stage I: Visual Manipulation}
The goal of this foundational stage is to instill robust visual synthesis and editing skills for mathematical diagrams. We pretrain the model on a mixture of our 5.2M-trajectory \editname and 10M-pair \imagenname datasets. To foster iterative editing capabilities, each editing trajectory is structured as a continuous sequence of 2-4 diagram transformations. To preserve the model's inherent reasoning abilities, we freeze the entire understanding pathway and exclusively train the Generation Expert via a Rectified-Flow Loss~\citep{liu2022flowstraightfastlearning} on the diagram generation task (Figure~\ref{fig:model}, Stage I). This approach builds a strong visual foundation without catastrophic forgetting of its core understanding capabilities.

\paragraph{Stage II: Strategic Visual-Aided Reasoning}
With the visual foundation established, Stage II fine-tunes the model to intelligently interleave its drawing and reasoning faculties using our interleaved image-text dataset, \datasetname. To enable the model to strategically decide \textit{when} to draw, it is trained on a token prediction task. Following each text segment (marked by the \texttt{<im\_end>} token), the model must predict whether to generate the \texttt{<|vision\_start|>} token to initiate a drawing, or the \texttt{<|endoftext|>} token to conclude the response.

To inform \textit{how} the model draws and understands, we process input and output images differently. All images provided in the question are encoded into clean VAE and ViT tokens, serving as visual context. For images within the solution, which the model must generate, we additionally include noised VAE tokens to compute the Rectified-Flow Loss. Unlike Stage I, all model components are unfrozen and trained jointly (Figure~\ref{fig:model}, Stage II). To enhance generation quality, we also leverage the architecture's inherent dual Classifier-Free Guidance mechanism during inference. This orchestration stage culminates in a model that can autonomously generate diagrams as intermediate steps to solve complex problems. Detailed training hyperparameters are provided in Appendix~\ref{app:training_details}.

\section{Experiments}

\begin{table*}[t]
\centering
\resizebox{\textwidth}{!}{%
\begin{tabular}{l|c|cc|cccccccc}
\toprule
\multirow{2.5}{*}{\textbf{Model}} &
\multirow{1.5}{*}{\textbf{MathVista}} &
\multicolumn{2}{c|}{\textbf{MathVerse}} &
\multicolumn{8}{c}{\textbf{MathVision}} \\
\cmidrule(lr){3-4}
\cmidrule(lr){5-5}
\cmidrule(lr){6-12}

~ & (GPS) & (Text Dominant) & (Text Lite) & (test) & AnaG & Angle & Area & Len & SolG & Alg & Others   \\
\midrule

BAGEL              & 68.8 &  49.2 & 42.0 & 24.1 & 26.2 & 31.8 & 25.0 & 28.7 & 22.1 & 17.1 & 23.1   \\
\textbf{\modelname} & 79.3 & 65.4 & 59.9 & 32.9 & 48.8 & 49.1 & 35.2 & 37.9 & 31.2 & 30.1 & 27.9 \\
$\Delta$ & \up{10.5} & \up{16.2} & \up{17.9} & \up{8.8} & \up{22.6} & \up{17.3} & \up{10.2} & \up{9.2} & \up{9.1} & \up{13.0} & \up{4.8} \\
\bottomrule
\end{tabular}
}
\caption{Generalization performance of \modelname{} compared to its base model (BAGEL) on three multimodal math benchmarks. $\Delta$ indicates the absolute improvement. MathVision subject abbreviations: AnaG (Analytic Geometry), SolG (Solid Geometry), Alg (Algebra), Angle (Metric Geometry - Angle), Area (Metric Geometry - Area), and Len (Metric Geometry - Length).}
\vspace{-1mm}
\label{tab:lmm_math_benchmarks}
\end{table*}

\begin{table}[t]
\centering
\resizebox{\columnwidth}{!}{%
\begin{tabular}{l|cc}
\toprule
\multirow{2.5}{*}{\textbf{Model}} &
\multicolumn{2}{c}{\textbf{Overall}} \\
\cmidrule(lr){2-3}
~ & \textbf{Complete} & \textbf{Weighted} \\
\toprule
\textbf{\modelname}                   & \textbf{21.9} & \textbf{34.4} \\
\quad w/o \editname   & 19.8          & 32.0          \\
\qquad w/o \imagenname & 18.2          & 30.8          \\
\bottomrule
\end{tabular}%
}
\caption{Ablation study on the pre-training corpora. We report the performance drop after removing the editing data (w/o \editname) and the entire pre-training data (w/o \imagenname).}
\label{tab:abla_data}
\end{table}
\begin{table}[t]
\centering
\small
\begin{tabular}{l|cc}
\toprule
\multirow{2.5}{*}{\textbf{Model}} &
\multicolumn{2}{c}{\textbf{Overall}} \\
\cmidrule(lr){2-3}
~ & \textbf{Complete} & \textbf{Weighted} \\
\toprule
\textbf{\modelname}     & \textbf{21.9} & \textbf{34.4} \\
-- (Skip Image)         & 19.7          & 31.9          \\
\modelname-Text         & 18.7          & 30.9          \\
\bottomrule
\end{tabular}%
\caption{Ablation study on the visual modality. \modelname-Text is a variant fine-tuned without any visual data. (-- Skip Image) denotes the full model being constrained to text-only reasoning during inference.}
\vspace{-1mm}
\label{tab:abla_image}
\end{table}

We compare \modelname{} against 20 prominent LMMs, including top-performing proprietary models such as the Gemini series (2.5-Pro, 2.5-Flash, Nano-Banana, 2.0-Flash)~\citep{comanici2025gemini25pushingfrontier}, the GPT series (GPT-5, GPT-4.1, GPT-4.1-mini, GPT-4o) \citep{openai2025gpt5systemcard,openai2024gpt4technicalreport,openai2024gpt4ocard}, Claude-Sonnet-4~\citep{anthropic_system_card_claude_4_2025}, other strong multimodal models like Seed-1.6-Thinking~\citep{seed2025seed15thinkingadvancingsuperbreasoning} and Qwen3-VL-Plus~\citep{Qwen2.5-VL}, and powerful open-source models, including the Qwen-2.5-VL series (7B, 32B, 72B)~\citep{Qwen2.5-VL}, Gemma-3-27b-it\citep{gemmateam2025gemma3technicalreport}
, InternVL3.5 (8B, 30B)~\citep{wang2025internvl35advancingopensourcemultimodal}, and Keye-VL-1.5-8B~\citep{yang2025kwaikeyevl15technical}. We also include our base model, BAGEL~\citep{deng2025emergingpropertiesunifiedmultimodal}, and a variant, BAGEL-Zebra-CoT~\citep{li2025zebra}, to precisely measure the gains from our framework. All LMM evaluations are conducted using VLMEvalKit~\cite{duan2024vlmevalkit} to ensure a fair comparison. The comprehensive results are shown in Table~\ref{tab:main_model_performance}.

\subsection{Benchmark Results}
As presented in Table~\ref{tab:main_model_performance}, \modelname{} achieves a weighted score of 34.4\% on our benchmark, establishing it as the top-performing open-source model. It surpasses all open-source competitors, including significantly larger models like Qwen-2.5-VL-72B (32.8) and InternVL3.5-30B-A3B (22.2). This result represents a substantial +15.9 point improvement over its base model, BAGEL, demonstrating the profound effectiveness of our training paradigm in unlocking advanced reasoning capabilities. Furthermore, \modelname{} proves to be highly competitive with proprietary systems, outperforming several prominent models such as Gemini-2.0-Flash (32.6) and GPT-4.1 (30.0).

An analysis of performance across mathematical domains reveals that \modelname{} exhibits the most significant gains in geometry-heavy subjects: Trigonometry (+27.1), Plane Geometry (+19.2), and Solid Geometry (+12.3). This result strongly supports our hypothesis that visual reasoning is particularly beneficial for geometric problem-solving. The model also shows substantial improvements in Analytic Geometry (+14.1) and Algebra (+11.8), suggesting that the ability to visualize functions and coordinate systems enhances reasoning in broader mathematical contexts. The modest gain in Calculus \& Vector (+0.8) indicates that this domain may require specialized reasoning capabilities beyond the scope of our current visual augmentation techniques.

\subsection{Performance on Other Math Benchmarks}
To assess the generalization capabilities of \modelname, we evaluate it on three established public benchmarks: the GPS category from MathVista's \texttt{testmini} set~\citep{lu2024mathvista}, the full \texttt{test} set of MathVision~\citep{wang2024measuring}, and the Text Dominant/Lite subsets from MathVerse's \texttt{testmini}~\citep{zhang2024mathverse}. As detailed in Table~\ref{tab:lmm_math_benchmarks}, \modelname{} demonstrates substantial and consistent improvements over its base model, BAGEL, across all benchmarks, with particularly strong gains on MathVerse (+17.9) and MathVista (+10.5). The detailed breakdown on MathVision further reveals significant improvements in subjects that benefit from visual intuition, such as Analytic Geometry (+22.6), Algebra (+13.0), and various plane geometry tasks (Angle: +17.3). Crucially, since these benchmarks require text-only solutions, this strong performance validates that our training paradigm fundamentally enhances the model's intrinsic reasoning abilities, allowing it to generalize effectively to traditional problem-solving formats.

\subsection{Ablation Studies}
We conduct a series of ablation studies to dissect the contributions of the key components within our framework: the pretraining corpus and the role of the visual modality in the final reasoning stage.

\paragraph{Effectiveness of the Pre-training Corpus.}
We investigate the impact of our two-stage pre-training strategy by ablating the \editname{} and \imagenname{} corpora. As shown in Table~\ref{tab:abla_data}, removing the \editname{} data (w/o \editname) results in a 2.4-point drop in the weighted score. This highlights the importance of learning step-by-step diagram editing, a critical skill for solving complex problems that require constructing auxiliary elements. A further ablation, removing the entire pre-training stage (w/o \imagenname), leads to an additional 1.2-point performance decrease. This confirms that even foundational diagram generation capabilities provide a vital scaffold for the fine-tuning phase. Together, these results validate our two-stage pre-training approach, demonstrating that both generation and editing skills are essential for achieving optimal performance.

\paragraph{Importance of Visual Modality in Reasoning.}
We analyze the importance of the visual modality through two ablations. First, we fine-tune a variant, \modelname-Text, using only the textual reasoning paths from \datasetname{}. Second, we constrain the full \modelname{} model to bypass visual generation during inference (-- Skip Image). As shown in Table~\ref{tab:abla_image}, both scenarios result in a significant performance drop. The \modelname-Text variant's weighted score falls by 3.5 points, confirming that training on interleaved visual-textual data is essential for learning complex reasoning. Interestingly, the model that simply skips image generation at inference (-- Skip Image) performs 1.0 point better than \modelname-Text, despite both producing text-only solutions. This suggests that our interleaved training paradigm not only teaches the model how to leverage visual aids but also fundamentally enhances its underlying textual reasoning capabilities.

\section{Conclusion}

We introduced \frameworkname, a comprehensive framework to endow Large Multimodal Models with intrinsic Visual Chain-of-Thought capabilities for mathematical reasoning. By leveraging our newly created large-scale datasets (\editname, \imagenname, and \datasetname) in a two-stage training recipe, we taught our model, \modelname, to master diagram manipulation and strategically interleave it with textual deduction. This approach yielded an 86\% relative improvement over strong baselines on our \benchmarkname benchmark. Crucially, this training paradigm not only teaches the model \textit{when} and \textit{how} to draw, but also fundamentally enhances its core textual reasoning. Our work provides a robust foundation for future research into broader and more complex multimodal reasoning.


\bibliography{custom}

\clearpage
\appendix

\section{Training Details}
\label{app:training_details}

We implement our framework on top of the publicly available \textbf{BAGEL-7B-MoT}~\citep{deng2025emergingpropertiesunifiedmultimodal} model. All training experiments were conducted on a cluster of 16 NVIDIA H800 GPUs. We use the AdamW~\citep{loshchilov2019decoupledweightdecayregularization} optimizer for both training stages. The detailed hyperparameters for our two-stage training recipe, corresponding to Stage I (Visual Manipulation) and Stage II (Strategic Visual-Aided Reasoning), are provided in Table~\ref{tab:hyperparams}.

\begin{table*}[h!]
\centering
\begin{tabular}{lcc}
\toprule
\textbf{Hyperparameter} & \textbf{Stage I} & \textbf{Stage II} \\
\midrule
\multicolumn{3}{l}{\textit{\textbf{Optimizer \& Scheduler}}} \\
Learning Rate (LR) & $2 \times 10^{-5}$ & $1 \times 10^{-5}$ \\
LR Scheduler & Cosine Decay & Cosine Decay \\
Min Learning Rate & $1 \times 10^{-7}$ & $1 \times 10^{-7}$ \\
Warmup Steps & 2,000 & 500 \\
Total Training Steps & 80,000 & 16,000 \\
\midrule
\multicolumn{3}{l}{\textit{\textbf{Model \& Loss}}} \\
EMA Decay Rate & 0.999 & 0.995 \\
Rectified-Flow Timestep Shift & 2.0 & 2.0 \\
Cross-Entropy (CE) Loss Weight & \texttt{N/A} & 0.25 \\
Rectified-Flow (MSE) Loss Weight & 1.0 (Implicit) & 1.0 \\
Frozen Components & Understanding Expert & None \\
\midrule
\multicolumn{3}{l}{\textit{\textbf{Batching \& Tokenization}}} \\
Max Tokens per Batch & 46,080 & 51,200 \\
Max Tokens per Sample & 8,192 & 25,600 \\
\midrule
\multicolumn{3}{l}{\textit{\textbf{Regularization (Dropout)}}} \\
Text Condition Dropout & 0.1 & 0.1 \\
ViT Condition Dropout & 0.3 & 0.1 \\
VAE Condition Dropout & 0.1 & 0.1 \\
\bottomrule
\end{tabular}%
\caption{Key hyperparameters for the two-stage training process. "N/A" indicates that the parameter was not applicable to that stage.}
\label{tab:hyperparams}
\end{table*}

\paragraph{Stage I} In this stage, the primary objective is to train the model's visual generation capabilities. As described in Section~\ref{sec:recipe}, we freeze the entire understanding expert and only train the generation expert. The loss is solely based on the Rectified-Flow objective~\citep{liu2022flowstraightfastlearning} for diagram generation, hence the absence of a Cross-Entropy loss component. We employed a slightly higher ViT condition dropout rate (0.3) to regularize the model and prevent overfitting to the visual features of the pretrainig data.

\paragraph{Stage II} In the second stage, all model components are unfrozen to enable joint optimization. The model is trained on a combined loss function: a Cross-Entropy loss for predicting the next token (either text or the special \texttt{<|vision\_start|>} and \texttt{<|endoftext|>} token), weighted by 0.25, and the Rectified-Flow loss for generating diagrams, weighted by 1.0. The learning rate is halved, and the number of training steps is reduced, which is typical for fine-tuning tasks. The ViT condition dropout is lowered to 0.1 to better leverage visual context during strategic reasoning.

\section{Dataset Details}
\label{app:dataset_details}

\subsection{\editname{} and \imagenname{}}
\label{app:pretrain_details}

\paragraph{Details on Foundational Structure Generation.}
As described in Section~\ref{sec:trainingdataset}, the Foundational Structure Generation pipeline for the MathCanvas-Edit dataset relies on an automated algorithm that randomly and incrementally adds geometric primitives and relations. This section specifies the exact sets used in this process.

\noindent\textbf{Geometric Primitive Set.} The generation process initiates by selecting one of 18 basic geometric objects from the following set:
\begin{itemize}[leftmargin=*, itemsep=0pt, topsep=2pt]
    \item \code{segment}, \code{angle}
    \item Triangles: \code{triangle}, \code{iso\_triangle} (isosceles), \code{r\_triangle} (right), \code{triangle_ab}, \code{ieq\_triangle} (equilateral), \code{risos} (right isosceles)
    \item Quadrangles: \code{rectangle}, \code{isquare}, \code{trapezoid}, \code{r\_trapezoid} (right), \code{eq\_trapezoid} (isosceles), \code{quadrangle},    \code{eq\_quadrangle} (equilateral), \code{eqdia\_quadrangle} (equal-diagonal)
    \item Polygons: \code{pentagon}, \code{eq\_pentagon} (equilateral)
\end{itemize}

\noindent\textbf{Geometric Relation Set.} Subsequently, the algorithm iteratively applies relations from a predefined set of 41 constructions. These are categorized by the number of new points they introduce (typically one or two).
\begin{itemize}[leftmargin=*, itemsep=0pt, topsep=2pt]
    \item \textbf{1-Point Relations (37):} \code{angle\_bisector}, \code{angle\_mirror}, \code{circle}, \code{circumcenter}, \code{eq\_triangle}, \code{eqangle2}, \code{eqdistance}, \code{foot}, \code{incenter}, \code{excenter}, \code{intersection\_cc}, \code{on\_bline}, \code{intersection\_lc}, \code{on\_aline}, \code{intersection\_ll}, \code{on\_line}, \code{intersection\_lp}, \code{intersection\_lt}, \code{intersection\_pp}, \code{intersection\_tt}, \code{lc\_tangent}, \code{midpoint}, \code{mirror}, \code{nsquare},  \code{on\_bline}, \code{on\_circle},  \code{on\_pline}, \code{on\_tline}, \code{on\_dia}, \code{orthocenter}, \code{parallelogram}, \code{psquare}, \code{reflect}, \code{s\_angle}, \code{shift},  \code{on\_opline}, \code{eqangle3}, \code{on\_circum}
    \item \textbf{2-Point Relations (4):} \code{square},  \code{trisegment}, \code{trisect}, \code{tangent}
\end{itemize}
The automated algorithm randomly samples from these sets to build progressively more complex diagrams, ensuring systematic coverage of fundamental geometric operations.

\begin{figure*}[t]
 \centering
 \includegraphics[width=1.0\linewidth]{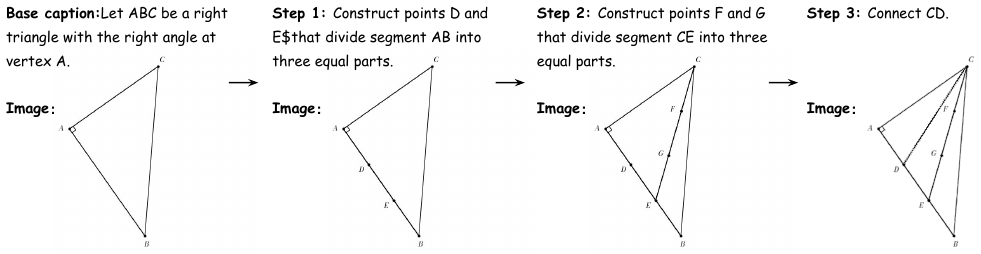}
 \caption{An example from MathCanvas-Edit dataset.}
 \label{fig:supp_edit_example1}
 \vspace{-1mm}
\end{figure*}

\begin{figure*}[t]
 \centering
 \includegraphics[width=1.0\linewidth]{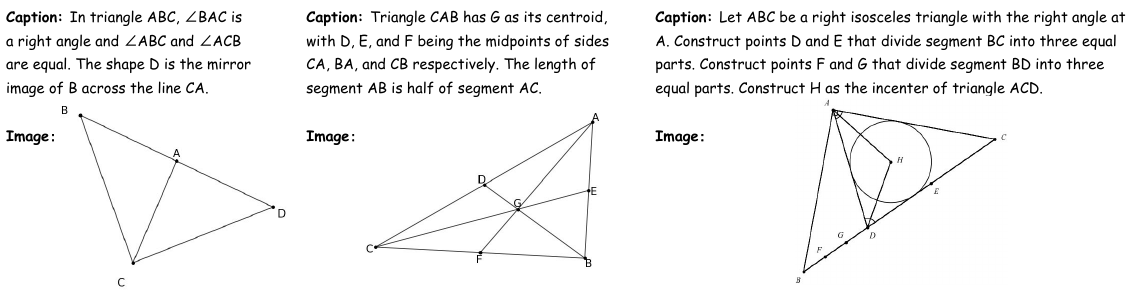}
 \caption{Examples from MathCanvas-Imagen dataset.}
 \label{fig:supp_t2i_example1}
 \vspace{-1mm}
\end{figure*}

\begin{figure*}[t]
 \centering
 \includegraphics[width=1.0\linewidth]{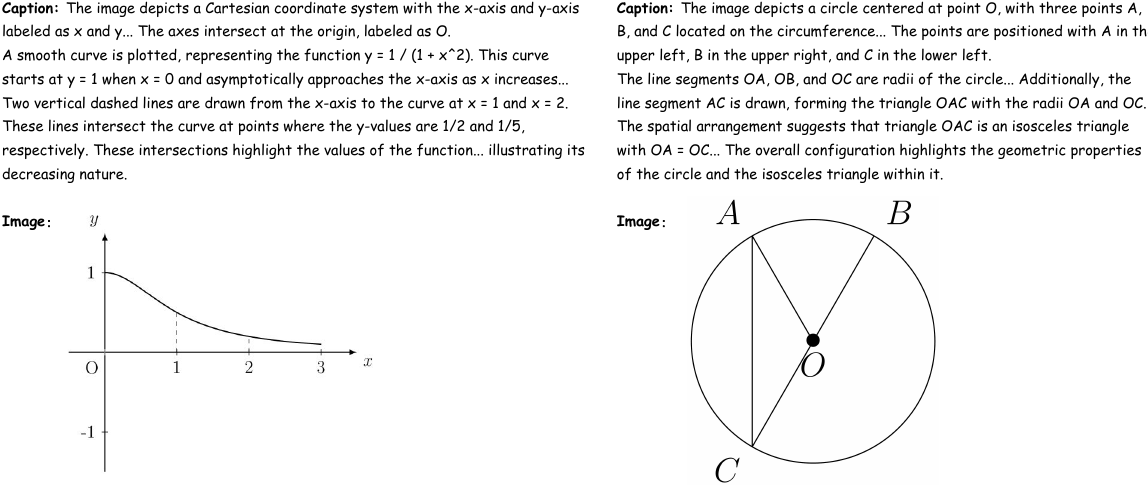}
 \caption{Examples from MathCanvas-Imagen dataset.}
 \label{fig:supp_t2i_example2}
 \vspace{-1mm}
\end{figure*}

\paragraph{Examples.}
An example from the \editname{} dataset is presented in Figure~\ref{fig:supp_edit_example1}. Examples from the \imagenname{} dataset are shown in Figures~\ref{fig:supp_t2i_example1} and \ref{fig:supp_t2i_example2}.

\subsection{\datasetname}
\label{app:sft_details}

\begin{figure}[t]
 \centering
 \includegraphics[width=1.0\linewidth]{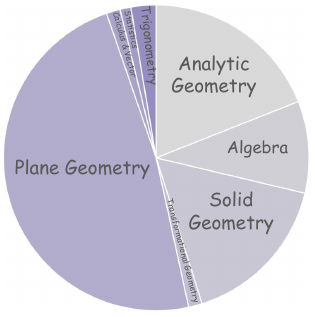}
 \caption{Distribution of knowledge type of MathCanvas-Instruct dataset.}
 \label{fig:supp_dataset}
 \vspace{-1mm}
\end{figure}

\paragraph{Dataset Statistics.}
We present the knowledge point distribution of the MathCanvas-Instruct training set in Figure~\ref{fig:supp_dataset}. Table~\ref{tab:supp_dataset_static} demonstrates the statistical characteristics of the MathCanvas-Instruct dataset, comprising 219K problems, of which 65\% are multimodal and 35\% are text-only. We have also analyzed the distribution of problem sources, the length of questions and solutions, and the number of images they contain.

\begin{table}[t]
    \centering 
    
    \begin{tabular}{lr}
        \toprule 
        \textbf{Statistics} & \textbf{Number} \\ 
        \midrule 
        
        \textbf{Total Samples} & \textbf{218,604} \\
        \midrule 
        
        - Text questions & 35\% \\ 
        - Multimodal questions & 65\% \\
        \midrule
        
        - Middle school questions & \textbf{63\%} \\
        \quad - Grade 7 & 6\% \\ 
        \quad - Grade 8 & 17\% \\
        \quad - Grade 9 & 77\% \\
        - High school questions & \textbf{37\%} \\
        \quad - Grade 10 & 12\% \\
        \quad - Grade 11 & 16\% \\
        \quad - Grade 12 & 72\% \\
        \midrule
        
         \quad - One question & 68\% \\
        \quad - Two sub-questions & 18\% \\
        \quad - Three sub-questions & 12\% \\
        \quad - Four or more sub-questions & 2\% \\
        \midrule
        
        \textbf{Question length (text tokens)} & \\ 
        - Maximum & 466 \\
        - Average & 107.92 \\
        \midrule
        
        \textbf{Solution length (text tokens)} & \\
        - Maximum & 2001 \\
        - Average & 539.66 \\
        \midrule
        
        \textbf{Multimodal Question Image} & \\
        - Maximum number & 5 \\
        - Average number & 1.03 \\
        \midrule
        
        \textbf{Solution Image} & \\
        - Maximum number & 5 \\
        - Average number & 1.18 \\
        
        \bottomrule 
    \end{tabular}
    
\caption{More statistics of MathCanvas-Instruct dataset.}
\label{tab:supp_dataset_static}
\end{table}

\begin{figure*}[t]
 \centering
 \includegraphics[width=1.0\linewidth]{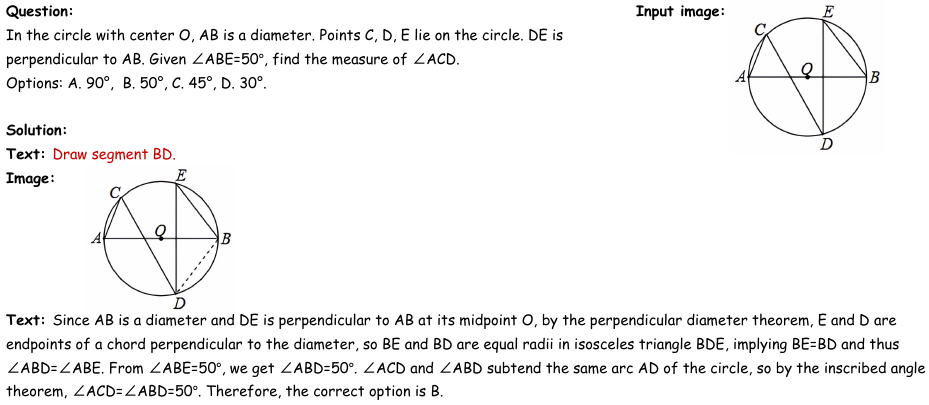}
 \caption{An example from MathCanvas-Instruct dataset.}
 \label{fig:supp_sft_example1}
 \vspace{-1mm}
\end{figure*}

\begin{figure*}[t]
 \centering
 \includegraphics[width=1.0\linewidth]{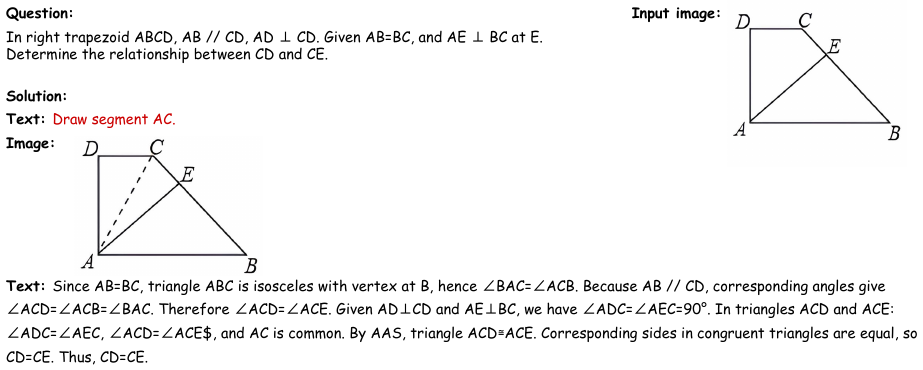}
 \caption{An example from MathCanvas-Instruct dataset.}
 \label{fig:supp_sft_example2}
 \vspace{-1mm}
\end{figure*}

\begin{figure*}[t]
 \centering
 \includegraphics[width=1.0\linewidth]{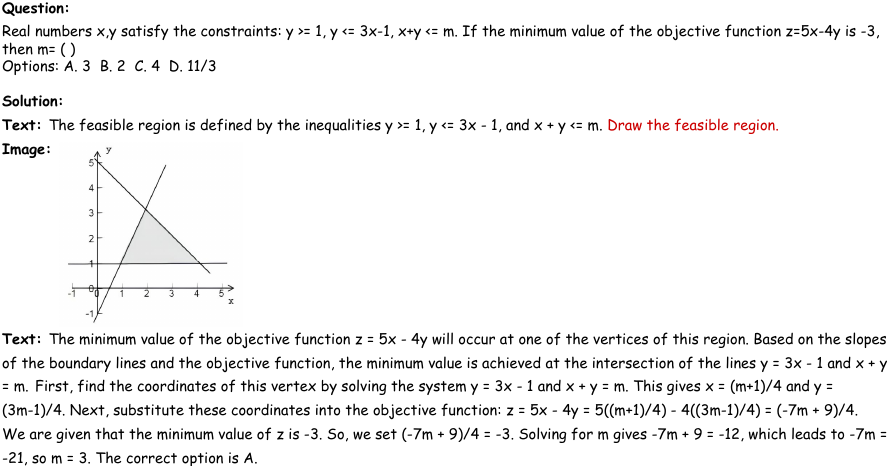}
 \caption{An example from MathCanvas-Instruct dataset.}
 \label{fig:supp_sft_example3}
 \vspace{-1mm}
\end{figure*}

\paragraph{Examples.}
We showcase examples from the \datasetname{} dataset in Figures~\ref{fig:supp_sft_example1}, \ref{fig:supp_sft_example2}, \ref{fig:supp_sft_example3}.

\section{Benchmark Evaluation Details}
\label{app:eval_details}

\subsection{Weighted Scoring Weights}
The weights for our Weighted Scoring metric are calculated using an exponential growth factor of 1.3, valuing later sub-questions more heavily. The specific formula for the weight $w_i$ of the $i$-th sub-question in a problem with $N$ sub-questions is:
\[
w_i = \frac{1.3^{i-1}}{\sum_{j=1}^{N} 1.3^{j-1}}
\]
Since our benchmark contains problems with a maximum of four sub-questions, we use the following pre-calculated, normalized weights for evaluation. The final score for a problem is the sum of the weights of the correctly answered sub-questions.

\begin{itemize}[leftmargin=*, itemsep=2pt]
    \item \textbf{For 2 sub-questions:} [0.4348, 0.5652]
    \item \textbf{For 3 sub-questions:} [0.2506, 0.3258, 0.4236]
    \item \textbf{For 4 sub-questions:} [0.1616, 0.2101, 0.2732, 0.3551]
\end{itemize}

\subsection{Evaluation Template}
Tables~\ref{tab:supp_prompt1} and \ref{tab:supp_prompt2} display the prompt templates used for MathCanvas-Bench evaluation.

\begin{table*}[!h]
\centering
\begin{tcolorbox}[colframe=black, colback=gray!5, arc=5mm, boxrule=0.5mm, width=\textwidth]
\begin{tabular}{p{\linewidth}}
You are an expert mathematics teacher and a precise data evaluator. Your task is to analyze a given math problem, compare a predicted solution against a ground truth answer, and determine if the prediction is correct.
\vspace{0.5em}

\textbf{INPUT FORMAT:} \\
You will be provided with a JSON string containing the following fields:
\begin{itemize}
\itemsep0em
\item \texttt{question\_text}: The full text of the mathematical problem.
\item \texttt{ground\_truth\_answer}: The correct, final answer text. This is the gold standard and is already extracted.
\item \texttt{prediction\_solution}: The full solution text from the model, from which you must extract the final answer(s).
\end{itemize}
\vspace{0.5em}

\textbf{TASK \& OUTPUT REQUIREMENTS:} \\
Your output must be a single, valid JSON object. The process involves two main steps: \textbf{Answer Parsing and Extraction} and \textbf{Correctness Judgment}.
\vspace{0.5em}

\textbf{Step 1: Answer Parsing and Extraction} \\
Your first task is to create two lists of answers: \texttt{gt\_answers} and \texttt{pred\_answers}. The structure of the \texttt{gt\_answers} list defines the required structure for the \texttt{pred\_answers} list.

\textbf{1.1 Parsing \texttt{ground\_truth\_answer}:}
\begin{itemize}
\itemsep0em
\item The \texttt{ground\_truth\_answer} is a clean, final answer.
\item Your task is to \textbf{parse} it into a list (\texttt{gt\_answers}).
\item \textbf{CRITICAL PARSING RULE:} The \textbf{only} condition for creating a list with multiple elements is the presence of explicit multi-part answer tags (e.g., \texttt{<1>...</1>}, \texttt{<2>...</2>}).
\item \textbf{If tags are present:} Extract the content of each tag into a separate list element. \textit{Example:} \texttt{"<1>5 cm</1><2>10 cm</2>"} becomes \texttt{["5 cm", "10 cm"]}.
\item \textbf{If no such tags are present:} The \textbf{entire, unmodified string} must be treated as the \textbf{single element} of the list. Do not split the string by characters, words, commas, or any other pattern. \textit{Example 1:} \texttt{"ABC"} must become \texttt{["ABC"]}, \textbf{NOT} \texttt{["A", "B", "C"]}. \textit{Example 2:} \texttt{"x=5, y=10"} must become \texttt{["x=5, y=10"]}, \textbf{NOT} \texttt{["x=5", "y=10"]}.
\item The \texttt{gt\_answers} list will never contain \texttt{null} elements and its length defines the number of sub-questions.
\end{itemize}

\textbf{1.2 Extracting from \texttt{prediction\_solution}:}
\begin{itemize}
\itemsep0em
\item Your primary task is to \textbf{extract} the final answer(s) from the \texttt{prediction\_solution} text to create the \texttt{pred\_answers} list.
\item \textbf{IMPORTANT}: The answers to different sub-questions may appear in different places within the \texttt{prediction\_solution}, not necessarily grouped together at the end. You must treat this as a \textbf{matching task}.
\end{itemize}

\end{tabular}
\end{tcolorbox}
\caption{
The prompt template (part 1) used by GPT-4.1 for mathematical reasoning evaluation.}
\label{tab:supp_prompt1}
\end{table*}

\begin{table*}[!h]
\centering
\begin{tcolorbox}[colframe=black, colback=gray!5, arc=5mm, boxrule=0.5mm, width=\textwidth]
\begin{tabular}{p{\linewidth}}

\begin{itemize}
\itemsep0em
\item For each part of the \texttt{gt\_answers} list, you must scan the \textbf{entire} \texttt{prediction\_solution} to find the corresponding predicted answer. Look for explicit labels (e.g., "(1)", "Part A"), final conclusions, boxed answers (e.g., \texttt{\textbackslash boxed\{...\}}), or statements that directly answer a part of the original question.
\item The final \texttt{pred\_answers} list \textbf{must have the exact same length as the \texttt{gt\_answers} list}.
\item For each sub-question, if you cannot find a corresponding answer in the \texttt{prediction\_solution}, you \textbf{must} use \texttt{null} as a placeholder in that position. \textbf{This rule is critical and applies in all cases where an answer is missing, including when the \texttt{prediction\_solution} appears incomplete or is truncated before all sub-questions are addressed.}
\end{itemize}

\textbf{CRITICAL RULE:} The final \texttt{gt\_answers} and \texttt{pred\_answers} lists \textbf{must} be of equal length. The number of parts in the \texttt{ground\_truth\_answer} dictates the required length for both lists.
\vspace{0.5em}

\textbf{Step 2: Correctness Judgment} \\
Your second task is to compare the \texttt{pred\_answers} list against the \texttt{gt\_answers} list, element by element.

\textbf{Judgment Rules:}
\begin{itemize}
\itemsep0em
\item \textbf{Numerical Equivalence:} Treat numbers as correct if they are mathematically equivalent (e.g., \texttt{5}, \texttt{5.0}, \texttt{10/2}). Allow for minor floating-point rounding differences.
\item \textbf{Textual Equivalence:} For text answers, judge based on semantic meaning, not exact matching. Ignore case, whitespace, and phrasing differences (e.g., "CB is parallel to PD" is equivalent to "Line CB || Line PD").
\item \textbf{Generate Correctness List:} Create a boolean list named \texttt{correctness}. The i-th element is \texttt{true} if the i-th predicted answer is correct, \texttt{false} otherwise. This list \textbf{must} have the same length as the answer lists.
\end{itemize}
\vspace{0.5em}

\textbf{Final JSON Output Structure:} \\
Your entire response must be a single, valid JSON object matching the schema below. Do not include any text outside of this JSON object.
\begin{verbatim}
{
  "analysis": "A brief step-by-step explanation...",
  "gt_answers": [
    "string",
    ...
  ],
  "pred_answers": [
    "string or null",
    ...
  ],
  "correctness": [
    true/false,
    ...
  ]
}
\end{verbatim}
\vspace{0.5em}

\textbf{INPUT DATA:} {\textcolor{cyan}{\{input\_data\}}}
\end{tabular}
\end{tcolorbox}
\caption{
The prompt template (part 2) used by GPT-4.1 for mathematical reasoning evaluation. The text highlighted in \textcolor{cyan}{cyan} is replaced with the specific input data for each problem being evaluated.}
\label{tab:supp_prompt2}
\end{table*}

\section{Additional Qualitative Results}
\label{app:reasoning_results}
\begin{figure*}[t]
 \centering
 \includegraphics[width=1.0\linewidth]{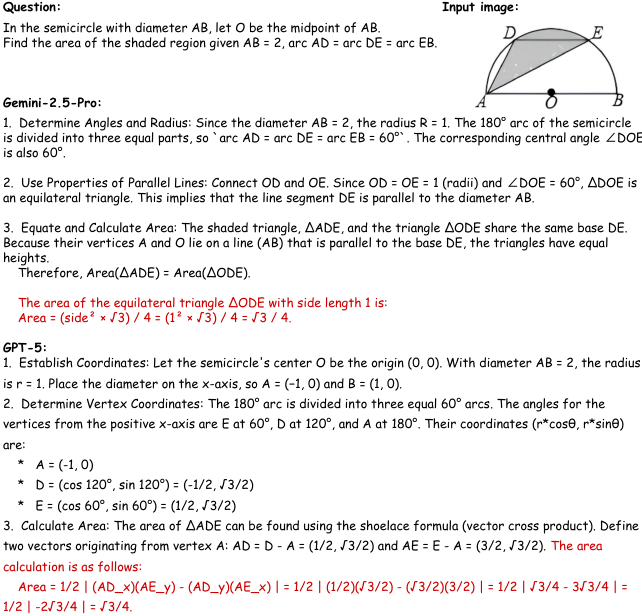}
 \caption{Leading LMMs (Gemini-2.5-Pro and GPT-5) solving the problem from Figure~\ref{fig:teaser} via text-only reasoning. Their complex or incorrect algebraic approaches, in contrast to the elegant geometric solutions unlocked by intrinsic VCoT, demonstrate the necessity of our \frameworkname{} framework.}
 \label{fig:teaser_LMM}
 \vspace{-1mm}
\end{figure*}

\begin{figure*}[t]
 \centering
 \includegraphics[width=1.0\linewidth]{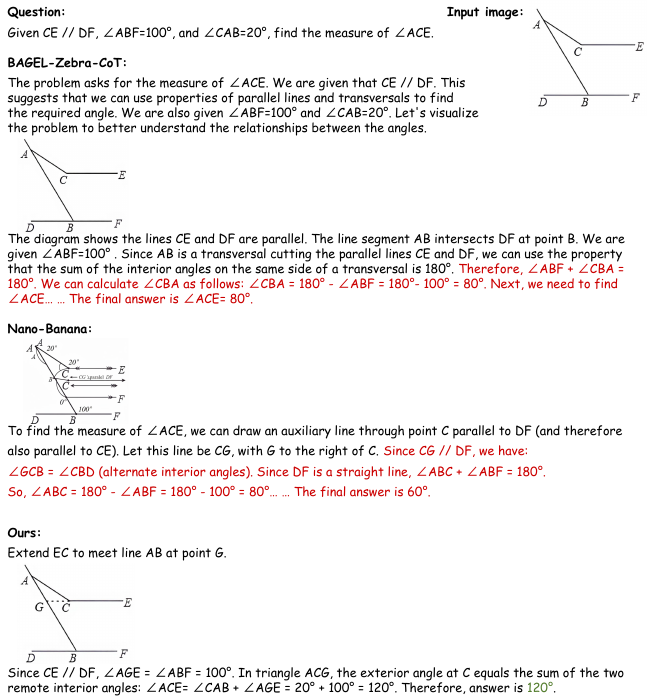}
 \caption{Comparison of BAGEL-Zebra-CoT, Nano-Banana, and our method.}
 \label{fig:response}
 \vspace{-1mm}
\end{figure*}

To further illustrate the limitations of even the most advanced LMMs when they lack intrinsic VCoT capabilities, we present qualitative examples of their performance on problems that benefit from visual manipulation. Figure~\ref{fig:teaser_LMM} shows the solutions from Gemini-2.5-Pro and GPT-5 for the problem featured in Figure~\ref{fig:teaser} of the main paper, demonstrating their reliance on complex and sometimes flawed algebraic approaches.
We provide more qualitative results of BAGEL-Zebra-CoT, Nano-Banana, and our method in Figure~\ref{fig:response}.

\end{document}